\documentclass[sigconf]{acmart}
\AtBeginDocument{%
  }

\setcopyright{acmlicensed}

\acmConference[Carrier Bag Narratives for More-than-Human Care Workshop]{Carrier Bag Narratives for More-than-Human Care Workshop}{June 14, 2026}{Singapore}
\acmBooktitle{Carrier Bag Narratives for More-than-Human Care Workshop), June 14, 2026, Singapore}

\acmISBN{978-1-4503-XXXX-X/2026/04}

\renewcommand\footnotetextcopyrightpermission[1]{}

\begin{document}
\title[From Social Imitation to the Social Becoming of Human Groups]
{A Robot Among People: \\From Social Imitation to the Social Becoming of Human Groups}

\author{Victor Tuan Vu Pham}
\orcid{0009-0007-2801-207X}
\affiliation{%
  \institution{Ubiquitous Design / Experience \& Interaction,\\ University of Siegen}
  \city{Siegen}
  \country{Germany}
}
\affiliation{%
  \institution{Honda Research Institute Europe}
  \city{Offenbach am Main}
  \country{Germany}
}
\email{tuan2.pham@uni-siegen.de}

\author{Judith Dörrenbächer}
\orcid{0000-0002-3968-461X}
\affiliation{%
  \institution{Ubiquitous Design / Experience \& Interaction,\\ University of Siegen}
  \city{Siegen}
  \country{Germany}
}
\email{judith.doerrenbaecher@uni-siegen.de}

\author{Thomas H. Weisswange}
\orcid{0000-0003-2119-6965}
\affiliation{%
  \institution{Honda Research Institute Europe}
  \city{Offenbach am Main}
  \country{Germany}
}
\email{thomas.weisswange@honda-ri.de}

\author{Marc Hassenzahl}
\orcid{0000-0001-9798-1762}
\affiliation{%
  \institution{Ubiquitous Design / Experience \& Interaction,\\ University of Siegen}
  \city{Siegen}
  \country{Germany}
}
\email{marc.hassenzahl@uni-siegen.de}

\renewcommand{\shortauthors}{Pham et al.}

\begin{abstract}
Robots designed to mediate human groups often fall into the solutionist trap: they are framed as sociable agents that fix problems such as conflict, disengagement, or lack of coordination. We suggest a different way of thinking. Rather than discrete agents, robots can be understood as situated elements of shared environments; catalysts and carriers of group experience whose meaning emerges through how people position, interpret, and interact with them. From this perspective, robots are not there to repair some ostensible dysfunctionality, but to enable group-level sense-making around care, norms, and identity. Our prior work on robotic street furniture suggests that this does not happen by imitating human sociality but by taking the shape of deliberately constrained, group-facing entities that happen and act for \textit{us} without being socially entangled as \textit{one of us}. We thus understand robots in public spaces not in terms of autonomy or intelligence, but as a relational capacity. This implies designing robots not in our image or for our utility, but grounded in our \textit{needs in being and becoming together}.
\end{abstract}

\maketitle

\section{Introduction}
Robots are increasingly envisioned as facilitators of human groups (see \cite{dahiya_survey_2022, schneiders_non-dyadic_2022, sebo_robots_2020, weisswange_design_2025} for recent reviews). This perspective understands robots as mediating agents that positively shape group dynamics and interpersonal relationships. Yet, these robots are rarely truly \textit{of} the groups they are meant to support. They speak politely, gesture appropriately, and show calibrated traces of emotion to mimic social competence. But when placed in real group situations and public places, they often remain peripheral, noticed and interacted with, sometimes even liked, but rarely woven into the dynamics that matter.

This is not because they fail technically. It is because they are framed as social actors in their own right: entities that enter a group and perform sociability (like talking, empathizing, mediating) while the group remains the background against which this performance is evaluated. Success is then measured by how individuals respond to the robot, rather than by whether its presence reshapes group interaction. Viewed like this, many robots function as \textbf{social props}, standing in for missing social labor. The issue begins with the assumption that they must simulate forms of human participation instead of participating \textit{differently}. 

As Hassenzahl and colleagues describe, robots (as well as other agentive systems) are experienced as \textit{Otherware}: not extensions of ourselves, but technological, considerable others with their own mode of presence \cite{hassenzahl_otherware_2021}. This otherness, grounded in their material and technical form, does not need to be overcome. Instead, robots can bring distinct “superpowers” into social situations (such as endless patience, non-competitiveness, or not taking things personally \cite{neuhaus_how_2022}). Neuhaus et al. recommend designers to consider such unique robotic properties rather than trying to mitigate or conceal them.

Correspondingly, by positioning the robot as an autonomous social actor, too much responsibility is placed on its performance and too little on the social situation it enters. Instead of asking how robots can more convincingly enact social skills, we propose understanding them as elements of  group experiences and shared spaces; carriers of implicit narratives; artifacts whose meaning emerges through collective sense-making. In this view, the robot becomes social only through how the group orients to it.

\section{The Hidden Assumption: Groups are Broken}
Much work on social and mediating robots starts from an assumption: that groups have problems in need of (technological) correction. Conflict needs to be reduced. Participation needs to be balanced. Discussions need to be moderated. Engagement needs to be increased. Once this assumption is in place, the robot’s role is clear. It becomes a physically instantiated tool for intervention, often introduced from the outside to nudge the group toward a predefined ideal of getting along socially.

But being together is not a broken state. Groups are not systems waiting to be fixed. They are ongoing, negotiated, value-laden processes, and emergent, less tangible climates \cite{forsyth_group_2010}. Tension, disagreement, care, and ambiguity are not failures of groups: inequality and friction are just as constitutive of group interaction as successful coordination and achieving shared understanding. Designing robots as solutions to social problems flattens this complexity and treats togetherness as something that can be optimized like an equation or a balance sheet.

\section{Not Social Actors but Group Experiences}
We propose a different starting point. Robots are physically instantiated elements that take up space and shape both the material and social configuration of a situation. More particularly, a robot in a human group does not carry a fixed social meaning of its own. Its role is not determined solely by its behavior, intelligence, or autonomy. Rather, its significance emerges through how the group positions it, talks about it, resists it, and incorporates it into their ongoing practices. In this sense, the robot is what the group makes of it; its behavior only begins to matter within the social space and context in which it appears (cf. \cite{barad_meeting_2007}). This also means that robots participate in shaping the narratives through which groups understand themselves.

Yet, many HRI experiments treat interaction with robots as a quasi-comprehension task, asking whether participants “correctly” understand a robot behavior or respond to it “as intended”. While such clarity may be important for more utilitarian purposes, such as usability or safety, it misunderstands the nature of social interaction.

\subsection{Relational Anchoring of Robot Meaning}
\begin{figure}[t]
    \centering
    \includegraphics[width=1\linewidth]{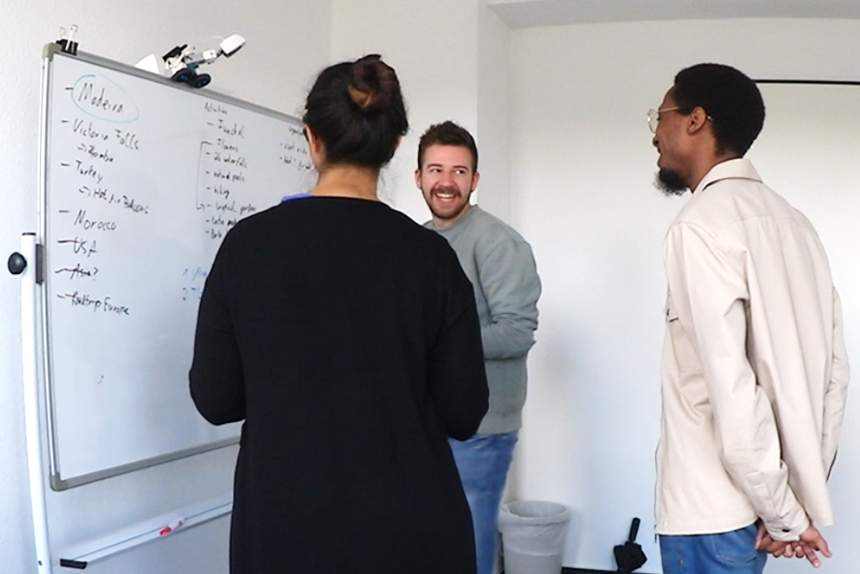}
    \caption{Lab group study with a small non-humanoid robot sitting on a whiteboard. The same robot behaviors, such as nodding and shaking of the head as well as gaze-following, elicited different interpretations of the robot, depending on who owned the robot and how the members got along.}
    \label{fig:labstudy}
\end{figure}

Across a series of experimental and speculative studies on human–robot group interaction \cite{pham_embodied_2024, pham_impact_2025, Doerrenbaecher2026MediatingUrbanSocialEncounters}, we approached robots less as agents but as situated vehicles of ongoing group dynamics. In controlled laboratory and video studies (see Fig. \ref{fig:labstudy}), we observed how people interpreted and responded to minimal robotic behaviors (nonverbal social cues like gaze and nodding from a non-humanoid designed as part of a whiteboard) during a group task. There, we found differences in participants' perceptions of a robot's legitimacy, alignment, and role in the group, as well as participants' judgments on the group and individual fellow members.

We then varied who the robot was owned by (i.e., an external entity, a speaking group member, the participant, or the group collectively) \cite{pham_who_2026}. Despite identical behavior, the robot was assigned different meanings: depending on how ownership positioned the robot in relation to the group and its individual members, the robot could be viewed as an ally, an outsider, a shared resource, the group's mouthpiece, or a subtle disruptor.

Ownership here did not simply impact how the robot was evaluated, but how the situation itself was understood. It shaped whose perspective the robot was seen to support and whether its behavior was interpreted as inclusive participation, neutral background activity, or strategic influence within the group.

\textit{Thus}, what changed was not the robot itself, but the collective orientation toward it and the interaction unfolding around it (cf. \cite{barad_diffracting_2014}). This does not hinge solely on the robot's behavior: before a robot speaks, gestures, or mediates, it is already socially positioned. Although easily backgrounded and often overlooked, subtle cues, such as who introduces it, who stands closest to it, and who is then assumed to control or be responsible for it, meaningfully anchor the robot within the group's relational fabric. The meaning of a robot is thus structured relationally (i.e., \textit{not only} behaviorally) and emerges through how the group makes sense of it.

\begin{figure}[t]
    \centering
    \includegraphics[width=1\linewidth]{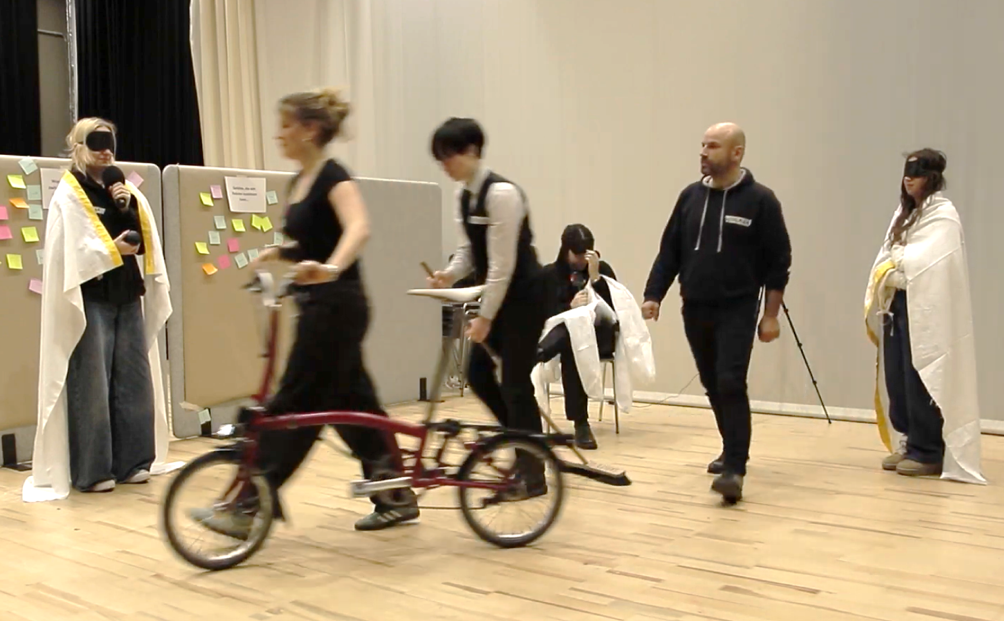}
    \caption{Theater-based co-design with adolescents wherein they ideated and enacted robots for multi-human scenarios in public space.}
    \label{fig:theaterworkshop}
\end{figure}

\subsection{Negotiating Robot Roles Through Enactment}
Complementing this experimental work, our theater-based co-design with adolescents \cite{Doerrenbaecher2026MediatingUrbanSocialEncounters} invited participants to enact robots in speculative public scenarios (Fig. \ref{fig:theaterworkshop}). In these workshops, robots were not approached as autonomous social actors, but as forms of \textit{interactive urban street furniture} situated within shared public spaces: elements that participants could position, appropriate, and enact as part of shaping social encounters. They occupied space, structured proximity, and made certain forms of interaction more or less likely. Here, too, the meaning of a robot emerged from the roles the groups of people assigned to it in situ. Adolescents used these robot-objects to stage situations in which they functioned as icebreakers, matchmakers, scapegoats, referees, or playful opponents, and repeatedly framed the same robot qualities (neutrality, shamelessness, persistence) as either strengths or weaknesses depending on the situation. This ambivalence supports our relational view of \textit{robotness}: what matters is not what the robot ``is'', but how its presence as a situated artifact reorganizes attention, accountability, and possibilities for being together in space.

Across both empirical and speculative contexts, roles were not fixed properties of the robot, but emergent outcomes of situated interpretation. 

We thus propose that ambiguity is not a flaw to be eliminated \cite{barad_diffracting_2014} through explanation, but a resource through which groups reveal (often implicit) norms, expectations, and values and make them speakable. Rather than asking whether people understand a robot’s behavior or purpose, studies should ask how that behavior reconfigures relationships, responsibilities, and the experience of being together.

In this way, robots are not social actors but elements of shared environments through which group experiences are composed. I.e., \textit{how} people perceive, live and articulate what matters to them as a collective. This shifts the design question from \textit{“What should the robot do?”} to \textit{“What kind of togetherness does this robot propose?”} and \textit{“What kinds of group maintenance does the robot make possible or visible?”}

\section{Designing for the “We”}
When people say that robots are human creations made “in our own image”, they usually mean that robots are made to look or act like humans. We argue the opposite. Anthropomorphism can short-circuit reflection by importing familiar social scripts too soon. Although this makes robots easier to describe and evaluate, it keeps them at a safe distance from what is actually at stake in groups.

We believe that robots should reflect us not in appearances or conduct, but in the social visions they materialize. A robot designed to calm people down expresses one understanding of collective being; a robot designed to surface shared concerns expresses another. These visions are normative, whether designers acknowledge them or not. Robots in groups are therefore not participants to be made more sociable nor tools meant to fix interpersonal problems. They are \textbf{designed propositions} that enter ongoing processes of collective sense-making. Their significance lies not in what they do on their own, but in how they reorganize the dynamics among the people around them. In other words, this is not about orchestrating isolated interactions, but about shaping the conditions under which group life unfolds. This includes how the relationships therein are initiated, transferred, repaired and transformed over time \cite{gillespie_rethinking_2014}.

Thus, the question for HCI and HRI is not how robots can better imitate social behavior, but how they can support groups in reflecting on who they are and how they want to be together. Designing such robots means taking responsibility for the social worlds their presence brings into being. When robot concepts become diffuse, then their most important role may not be to stand in for social group interaction, but to help people recognize, negotiate, and inhabit it collectively.

\section{Outlook}
We understand robots in groups not as corrective agents, but as relational propositions within ongoing processes of collective becoming. We thus suggest robot design in human group contexts to:
\begin{itemize}
    \item Act as relational conduits and catalysts of ongoing group dynamics
    \item Surface interpersonal and collective assumptions that would otherwise remain implicit
    \item Make tensions and differences addressable rather than suppressing them
    \item Create conditions for the renegotiation of norms and shared identity
\end{itemize}
This perspective shifts attention from robot behavior to group experience, and from optimization to responsibility. If robots manifest as an emergent relational capacity, then designing robots becomes inseparable from designing the conditions under which people \textit{are} and \textit{become together}.

\section*{Attending Authors}
\textbf{Victor Tuan Vu Pham} is a final-year PhD researcher in the Experience \& Interaction group at the University of Siegen (Germany). His research focuses on Human–Human–Robot Interaction, examining how non-humanoids (through nonverbal behavior, role attribution, and relational positioning) can influence socio-emotional aspects of human groups, such as experienced cohesion, belonging, and interpersonal perception. His work spans controlled lab studies and speculative approaches, including research on co-designing robot mediators for social encounters in public spaces.

\textbf{Judith Dörrenbächer} is a postdoctoral researcher at the University of Siegen (Germany) whose work explores speculative and performative methods in design, techno-animism, and more-than-human perspectives. She investigates how speculative design can open alternative imaginaries, with a particular focus on social robots. Judith is co-editor of "Meaningful Futures with Robots", published widely on robots and speculative design.
\bibliographystyle{ACM-Reference-Format}
\bibliography{bibliography.bib}

\end{document}